\documentclass[sigconf]{acmart} 

\usepackage{amsmath,amsfonts,bm}

\def\eqref#1{equation~\ref{#1}}

\def\1{\bm{1}}

\DeclareMathAlphabet{\mathsfit}{\encodingdefault}{\sfdefault}{m}{sl}
\SetMathAlphabet{\mathsfit}{bold}{\encodingdefault}{\sfdefault}{bx}{n}

\usepackage[utf8]{inputenc} 
\usepackage[T1]{fontenc}    
\usepackage{hyperref}        
\usepackage{url}            
\usepackage{booktabs}       
\usepackage{amsfonts}       
\usepackage{nicefrac}       
\usepackage{microtype}      
\usepackage{amsmath}
\usepackage{amssymb}
\usepackage{color}
\usepackage{url}
\usepackage{graphicx}
\usepackage{enumitem}
\usepackage[all]{nowidow}
\usepackage{comment}
\usepackage{booktabs}
\usepackage{mathrsfs}
\usepackage{algorithm}
\usepackage{algorithmic}
\usepackage{hyperref}
\usepackage{url}
\usepackage{wrapfig}
\usepackage{amsthm}
\usepackage{subcaption} 
 
\author{Ugo Tanielian}
\affiliation{%
  \institution{Criteo Research}}
\affiliation{%
  \institution{UPMC}
  \city{Paris} 
  \state{France}}
\email{u.tanielian@criteo.com}
\author{Flavian Vasile}
\affiliation{%
  \institution{Criteo Research}
  \city{Paris}
  \state{France} 
  \postcode{43017-6221}}
\email{f.vasile@criteo.com}

\setcopyright{acmlicensed}
\begin{document}

\title{Relaxed Softmax for learning from Positive and Unlabeled data}

\begin{abstract}
  In recent years, the softmax model and its fast approximations have become the de-facto loss functions for deep neural networks when dealing with multi-class prediction. This loss has been extended to language modeling and recommendation, two fields that fall into the framework of learning from Positive and Unlabeled data.
  
  In this paper, we stress the different drawbacks of the current family of softmax losses and sampling schemes when applied in a Positive and Unlabeled learning setup. We propose both a Relaxed Softmax loss (RS) and a new negative sampling scheme based on Boltzmann formulation. We show that the new training objective is better suited for the tasks of density estimation, item similarity and next-event prediction by driving uplifts in performance on textual and recommendation datasets against classical softmax.
\end{abstract}



\copyrightyear{2019}
\acmYear{2019}
\acmConference[RecSys '19]{Thirteenth ACM Conference on Recommender Systems}{September 16--20, 2019}{Copenhagen, Denmark}
\acmBooktitle{Thirteenth ACM Conference on Recommender Systems (RecSys '19), September 16--20, 2019, Copenhagen, Denmark}
\acmPrice{15.00}
\acmDOI{10.1145/3298689.3347034}
\acmISBN{978-1-4503-6243-6/19/09}

\maketitle
section{Introduction}


One-class learning is a well-known paradigm where one tries to identify objects of a specific class amongst all objects. Its applications are numerous in language modeling and recommendation, since in many occasions negative data is either too expensive to obtain or too difficult to define. Depending on the availability of training data, we make the distinction between learning from \textbf{positive-only data} and learning from \textbf{positive and unlabeled data}. In the positive-only case, one can use the softmax cross-entropy loss and train a neural model to fit a target density. The softmax cross-entropy loss is a density estimation tool that allows to fit the distribution output by the model to the empirical distribution of the data. However, in recommendation and language modeling, many tasks fall into the framework of learning to rank. Indeed, when recommending movies or products from a user history, one is mainly interested in predicting correctly the top items only. 

Contrary to the positive-only case, Positive and Unlabeled (PU) learning ~\cite{elkan2008learning, li2003learning, lee2003learning, ren2014positive, paquet2013one}, leverages some non-labeled data to further improve the performance of the model in classification tasks. One of the classical methods is to identify possible \textit{negatives} in this unlabeled set. In this setting, one can define a \textit{negative distribution} and move the problem back to ordinary supervised learning. The authors of \cite{w2v} applied this technique to Natural Language Processing (NLP) for the training of the proposed Word2Vec architecture. Indeed, in this scheme, the neural network learns how to discriminate between positive pairs and non-observed sampled pairs using a standard binary-cross entropy loss. The authors argue that this training loss does not try to approximate softmax but rather learn high-quality word embeddings. 

In this paper, we argue that when dealing with language modeling or recommendation tasks, negative sampling schemes should move away from approximating the softmax function. Moving in this direction, we define a new \textit{Relaxed Softmax} loss by replacing the normalization constant in the softmax cross-entropy function with one computed over a subset of chosen \textit{negatives}. Similarly to PU learning, one defines a prior on the distribution of the negative data and use a penalization scheme to refine the decision boundary. This prior on potential negative data can also be seen as a regularization term and leads to better generalization properties. Obviously, as in PU learning, the choice of this \textit{negative distribution} largely impacts the performance of the model. 

In NLP, previous generic sampling techniques such as uniform sampling or popularity sampling treat negatives independently from the context item and may be doing too simple assumptions on the set of \textit{negatives}. Authors in \cite{yang2018breaking} recently underlined what they called the \textit{softmax bottleneck} and showed that softmax does not have enough capacity to model natural language. Indeed, in the case where the underlying data is highly context-dependent, they brought significant improvements by over-parametrizing the model with a mixture of softmaxes. Therefore in this paper, we argue that for applications such as language modeling and recommender systems, negative sampling distributions should be contextual. 

We also make the link with previous works in active learning, where one is interested in sampling the most \textit{informative} negatives in a set of unlabeled data. In our specific case, we will never access the label of the queried sample, but we advocate that some samples can be more informative than others. On the one hand, sampling data points that are very unlikely to be positive can be inefficient as they would not bring major changes in the model's parameters. On the other hand, sampling data points close to the decision boundary where the confidence in the labeling is low might lead to a deteriorated performance. In this paper, we propose a sampling scheme based on a new Boltzmann formulation and highlight the existence of an optimal negative sampling distribution. 


Overall, the main contributions of this paper are the following:
\begin{itemize}
    \item We propose a new \textit{Relaxed Softmax} (RS) loss for training neural networks in a PU setting. We show its regularization properties on synthetic datasets and its improvements in performance over the softmax cross-entropy training loss and the sampled softmax loss in the context of language modeling and recommendation.
    \item From the gradient analysis of the \textit{Relaxed Softmax} loss, we derive a new context-based negative \textit{Boltzmann Sampling}.
    \item In the newly proposed \textit{Boltzmann Sampling}, one can fine tune the negative distribution to a specific task. Doing so, we show significant uplifts on tasks such as density estimation, next-item prediction and word similarities.
\end{itemize}

We discuss related work on negative sampling schemes for softmax and previous work related to PU learning and active learning in Section~\ref{section:related_work} of this paper. In Section~\ref{section:approach} we formally introduce both the \textit{Relaxed Softmax} training and the \textit{Boltzmann Sampling} scheme. We highlight the performance of our method in Section \ref{section:experiments}, and conclude with ideas and directions for future work.

\section{Related Work}\label{section:related_work}

\subsection{Training neural language models}


Authors in \cite{bengio2003neural} were the first to define the training of a neural probabilistic language models using a softmax cross-entropy training loss. Softmax being computationally slow, fast approximations were proposed to speed up the training. Sampled Softmax, introduced by authors in \cite{IS}, showed both huge improvements in time complexity and state-of-the-art results \cite{adaptative_IS, JeanCMB14}. Sampled softmax avoids computing scores for every word in the vocabulary as one chooses a proposal distribution from which it is cheap to sample, and performs a biased importance sampling approximation of softmax's gradient. Noise Contrastive Estimation \cite{NCE_loss} was also introduced as an unbiased estimator of the softmax and was shown to be efficient for learning word embeddings~\cite{mnih2012fast}. On the contrary, the authors of ~\cite{w2v, mikolov2013efficient} proposed a \textit{negative sampling} loss whose goal is not to approximate softmax but to learn high-quality vector representations for words. They consequently train a classifier that can discriminative between positive pairs coming from the true distribution and fake ones sampled from a negative sampling distribution. 

The introduction of the previously mentioned negative sampling schemes, along with the Word2Vec architecture proposed by~\cite{w2v, mikolov2013efficient}, has enabled state-of-the-art results in terms of the quality of the learned word embeddings. Word2Vec has become a widely used architecture, having found use in applications such as document classification~\cite{le2014distributed}, user modeling~\cite{vasile2016meta}, and graph embedding~\cite{node2vec, narayanan2016subgraph2vec}. Recent work has led to further improvements in words embedding models~\cite{BojanowskiGJM16, pennington2014glove}; a notable example is the FastText model~\cite{JoulinGBM16}, which leverages n-gram features to improve word embedding quality.
 
\subsection{Negative sampling from a PU learning perspective}
Positive and Unlabeled (PU) learning was originally defined in~\cite{li2005learning}. It has been shown that applying PU learning to language modeling and recommendation can lead to improved performances. Authors in \cite{liu2002partially} were the first to apply the framework of PU learning to the classification of text documents and showed significant uplifts in accuracy. For the training of word embeddings, authors of \cite{swivel} define an heuristic on all unobserved pairs and encourage the model to respect a pairwise-dependent upper-bound on its PMI estimate, making use of greater information. In the field of collaborative filtering, authors of \cite{liang2016modeling} incorporate a model of user exposure to items such that each unobserved pair is treated differently. Interestingly, negative sampling schemes introduced in the previous subsection can be seen as a way of moving the problem to the PU learning case. Contrary to conventional generic sampling losses such as \textit{Sampled Softmax} or \textit{Noise Contrastive Estimation}, we advocate that negative sampling schemes should not try to approximate the softmax function but rather act as a way to induce an informative bias on the model of negatives. To this end, authors in~\cite{chen2018improving} provide an insightful analysis of negative sampling and show that negative samples that have high inner product scores with the context word provide more informative gradients. PU learning approaches have already shown some successes in the context of Recommender Systems as shown in \cite{liang2016modeling}. Also, as several previous works \cite{blei2003latent, mikolov2012context, yang2018breaking} stressed that language modeling is highly context-dependent, we advocate in this paper for a context-based negative sampling scheme.

\subsection{An active learning based negative sampling scheme}
In the case of active learning, the model has accessed to a small labeled dataset and a bigger pool of unlabeled data. The goal then is to query the labels of the most valuable data points among this extensive unlabeled set. Therefore, using a sampling method, one can extend the training set with wisely chosen points that reduce the generalization error made by the model (\cite{roy2001toward}) or the variance over the dataset (\cite{yu2006active}). In active learning, a famous area of research is uncertainty sampling (\cite{huang2015efficient, tong2001support}). In this setting, one queries the least certain instances to help the model refine the decision boundary. On the opposite, in maximum model change sampling, one is looking to sample points with highest norm gradients to increase the performance (\cite{bachman2017learning, kading2016large}). We refer the interested reader to \cite{yang2018benchmark} for a more exhaustive comparison of active learning approaches. In our case, we will never have access to the true label of the sampled points, but we argue that an ideal negative sampling should be a relative trade-off between sampling data points close to the decision boundary and sampling very unlikely data.

\section{Next-Item Prediction}\label{section:approach}

We will start this section by formally defining the task. Then, we will present the \textit{Relaxed Softmax} loss and will motivate the efficiency of this loss. Finally, we will discuss the properties of an ideal negative sampling distribution and show its efficiency on a toy dataset. 

\subsection{Framework}

\subsubsection{Notations}
Let $I$ and $J$ denote two sets of objects. The set $I$ will be called the \textit{context set} while the set $J$ will be denoted the \textit{target set}. This setting can be applied to language processing, user activity modeling and graph theory (respectively next word, event and node prediction). $I$ and $J$ can also be identical.

In the one-class setting applied to learning pairwise co-occurences, there exists an unknown generating process $P$ from which we can only observe samples gathered in a dataset $D^\star = \{ (i_1, j_1),...,(i_n, j_n) \} \in (I \times J)^n$. For each context item $i \in I$, $P_i$ denotes the conditional process on $J$ given the context item $i$. Similarly, for a context $i \in I$, we have access to limited number of observations $\{j_1, ... j_n \}$ conditionally to $i$ and we note $\hat{P_i}$ the conditional empirical distribution on $J$. In the task of next-item prediction, the goal is to learn the true conditional process $P_i$ and we will compare performances in Section \ref{section:experiments} on both density estimation and ranking metrics.

$\mathscr G=\{G^{\theta}\}_{\theta \in \Theta}$ denotes the family of functions, with parameters $\Theta \subset \mathbb{R}^p$. The model $G^\theta$ takes an input in $I \times J$ and outputs a score $G^\theta(i,j) \in \mathbb{R}$. In this paper, we use the same architecture as the Word2Vec model proposed by \cite{w2v, mikolov2013efficient} but change both the training loss and the negative sampling distribution. This model is composed of an embedding matrix $W$ and an output layer $O$. For a given pair context/target $(i,j) \in I \times J$, we have $G^\theta(i,j) := \langle W_i \cdot O_j \rangle $, where $W_i, O_j$ corresponds to the $i$th and $j$th row of respectively $W$ and $O$ and $<.>$ refers to the dot product operator. 

\subsubsection{Maximizing Likelihood Estimation}
Usually, one tries to solve the one-class problem by performing density estimation of the empirical distribution with Maximum Likelihood Estimation (MLE). For a given context item $i \in I$ and any parametrized model $G^\theta$, one defines an estimated conditional distribution $g_i^\theta$ on the set of target items $J$ using the softmax function as follows :
\begin{equation}\label{softmax}
    \forall \textbf{i} \in I, \ g_i^\theta(j) = \frac{ \exp(G^\theta(i,j))}{\exp(G^\theta(i,j)) + \sum_{j' \in J \setminus j} \exp(G^\theta(i,j'))}
\end{equation} 
Therefore, for a given pair $(i,j) \sim P$, minimizing the negative log-likelihood leads to the following MLE objective:
\begin{equation}\label{eq:softmax_loss}
L_{(i,j)} = - \ln \ g_i^\theta(j) = - \ G^\theta(i,j) + \ln \sum_{j' \in J} \exp(G^\theta(i,j'))
\end{equation}
Doing so, one models the conditional process $P_i$ as a categorical distribution. From a density estimation point of view, minimizing Eq.~\ref{eq:softmax_loss} means minimizing the Kullback-Leibler divergence between the empirical conditional distribution $\hat{P_i}$ and the modeled distribution $g_i^\theta$. When deriving the gradient, one has : 
\begin{align}\label{eq:softmax_gradient}
    \nabla_\theta L_{(i,j)} &= - \nabla_\theta G^\theta(i,j) +  \mathbb{E}_{j' \sim g_i^\theta} \nabla_\theta G^\theta(i,j') \\
    \mathbb{E} \nabla_\theta L_{(i,j)} &= - \mathbb{E}_{j \sim \hat{P_i}} \nabla_\theta G^\theta(i,j) +  \mathbb{E}_{j \sim g_i^\theta} \nabla_\theta G^\theta(i,j) \label{independance_assumption}
\end{align}
where $\hat{P_i}$ refers to the empirical true distribution. Eq~\ref{independance_assumption} assumes that all data points $\{(i,j_1), ..., (i,j_n) \}$ are i.i.d. sampled and that therefore the conditional generating process is a distribution on $\mathcal{P}(J)$. As mentioned in the previous section, the \textit{Sampled Softmax} loss tries approximating $\mathbb{E}_{j \sim g_i^\theta} \nabla_\theta G^\theta(i,j)$ by performing a biased importance sampling with a cheap-to-sample proposal distribution. 

In the next subsection, we stress the different drawbacks of optimizing MLE in the specific setting of learning from pair co-occurences in language modeling or recommendation. 

\subsection{Relaxing Softmax}

\subsubsection{Drawbacks of the MLE training scheme}
First, when creating a pair dataset from a text corpora, one links a context word $i \in I$ to several target items $\{j_1, ..., j_n \} \in J^n$ depending on the \textit{sliding window} parameter and add the following pairs $(i,j_1),..., (i,j_n)$ to the dataset. Conditionally to the context item $i$, the target items $\{j_1, ..., j_n \}$ are therefore not i.i.d. sampled from a categorical distribution. The independence assumption made when performing Maximum Likelihood Estimation is therefore not valid. In language modeling and recommendation, this can be corrected with architectures taking into account the impact of long range dependencies, such as recurrent and attention-based deep learning models \cite{le2012measuring, trinh2018learning, vaswani2017attention}. In addition to this, as suggested by authors of \cite{blei2003latent, yang2018breaking}the assumption that target items are identically distributed conditionally to a context item might not be valid. However, correcting this drawback would require a different parametrization of the model and is not a problem tackled in this paper.

Also, one of the drawbacks of the softmax cross-entropy is that it does not leverage any potential \textit{negative} pairs of items in $I \times J$. We advocate that one should move away from this method and bring the problem to PU learning. In this setting, one method is to make the assumption of possible \textit{negative data} in the set of unlabeled items. One defines a \textit{negative distribution} and then performs ordinary supervised classification \cite{li2005learning}. Approximating softmax would mean sampling negative items with a uniform distribution and therefore making too simple assumptions on the set of negatives. Instead, we advocate that one should define a more complex sampling distribution that fits the negatives. Consequently, we propose a dynamical and contextual negative sampling scheme. 

Finally, many tasks in language modeling and recommender systems do not strictly fall into density estimation but are rather ranking metrics. In language modeling, similarity and analogy metrics focus on the K-Nearest-Neighbours in the embedding space. Besides, when tackling the task of next-item prediction, many metrics such as Mean Percentile Rank or Precision (both metrics will be detailed later in Section \ref{section:experiments}) are ranking metrics and using a density estimator as MLE might not be an efficient tool.

\subsubsection{The Relaxed Softmax loss}
Modeling $P_i$ as a categorical distribution, one assumes all target items to be conditionally dependent, and as discussed previously, this might lead to poor results. To correct this, we relax the constraint on the conditional probabilities by replacing the normalization constant used in the full-softmax formulation with one computed over a subset of chosen items. The chosen negatives are obtained with i.i.d. samples in a chosen \textit{negative sampling distribution}.

Therefore, for a given context/target pair $(i,j) \sim P$ and a chosen negative sampling distribution $Q_i \subset J$, we define the following \textit{Relaxed Softmax} (RL) loss with the following equation:
\begin{equation}\label{eq:negatively_sampled_softmax}
    L_{(i,j)} = - G^\theta(i,j) + \ln \sum_{j' \in V_{(i,j)}} \exp(G^\theta(i,j'))
\end{equation}
where $V_{(i,j)} = \{(i,j_1),...,(i,j_n) \}, \ \ \forall k \in [1,n], j_k \sim Q_i$.

The \textit{Relaxed Softmax} can be considered as a general framework as it is defined independently of both the negative sampling distribution $Q_i$ and the number of negative samples $n$. Using this training loss, one moves the problem to PU learning and forces the model to refine its decision boundary between the positive distribution and the chosen negative distribution. Indeed, the chosen \textit{Relaxed Softmax} loss directly compares the scores of positive pairs coming from $P$ with negative pairs sampled from $Q_i$. At each step, this training loss will therefore penalize the target items sampled by the negative distribution. In the specific case where $Q_i$ is the uniform distribution, the \textit{Relaxed Softmax} is a Monte-Carlo approximation of the softmax function. In the next subsection, we now analyze the gradient and consistency properties of this loss.

\subsubsection{Gradient Analysis}
The sampled negatives impact the performance of the results and play a significant role in the direction of the gradient. Deriving the loss $L_{(i,j)}$ with respect to $\theta$, we have:
\begin{equation}
    \nabla_\theta L_{(i,j)} = - \nabla_\theta G^\theta(i,j) +\sum_{j' \in V_{(i,j)}} \hat{g}_i^\theta(j') \times \nabla_\theta G^\theta(i,j')
\end{equation}
where $\hat{g}_i^\theta(j) := \frac{\exp G^\theta(i,j)}{\sum_{j' \in V} \exp G^\theta(i,j')}$.

The gradient can be decomposed as two parts: the left maximizes likelihood of observing positive data whereas the right part penalizes pairs that were sampled under the negative distribution $Q_i$. We see that if a negative $j \sim Q_i$ is too far away from the decision boundary and too easily classified by the model, the magnitude of the gradient $\nabla_\theta G^\theta(i,j')$ w.r.t. this specific input will be small. In language modeling and recommendation, where there are significant differences among classes, sampling closer to the decision boundary of the model can help improving the model \cite{chen2018improving}. 

Interestingly, when the number of negatives sampled goes to $\infty$, the gradient has the following formulation:
\begin{equation}\label{eq:consistency_grad_NSP_loss}
    \nabla_\theta L_{(i,j)} = - \nabla_\theta G^\theta(i,j) + \mathbb{E}_{j \sim B} \nabla_\theta G^\theta(i,j)
\end{equation}
where $\forall j' \in J, B(j') = \frac{Q_i(j') \times \exp(G^\theta(i,j'))}{\sum Q_i \times \exp(G^\theta)}$. 

Comparing with MLE's gradient from Eq. \ref{eq:softmax_gradient}, the difference is that we replace $\mathbb{E}_{j \sim g_i^\theta}$ $\nabla_\theta G^\theta(i,j)$ with $\mathbb{E}_{j \sim B} \nabla_\theta G^\theta(i,j)$. The expectancy of the gradient is consequently not computed over the estimated density $g_i^\theta$ but rather a Boltzmann distribution re-weighted by some \textit{negative} distribution $Q_i$. It is therefore natural to approximate this gradient with a sampling scheme defined with \textbf{a Boltzmann distribution}. We present this new sampling in the next subsection.

\subsection{A Boltzmann Negative Sampling distribution}

\begin{figure*} 
    \centering
    \subcaptionbox{Negatives sampled with $B(E=G^\theta, D_i=1/P_i, T=0.25)$. Very likely data is sampled; it leads to a poor estimation. \label{low_temp}}
    {   
        \includegraphics[width=0.31\linewidth]{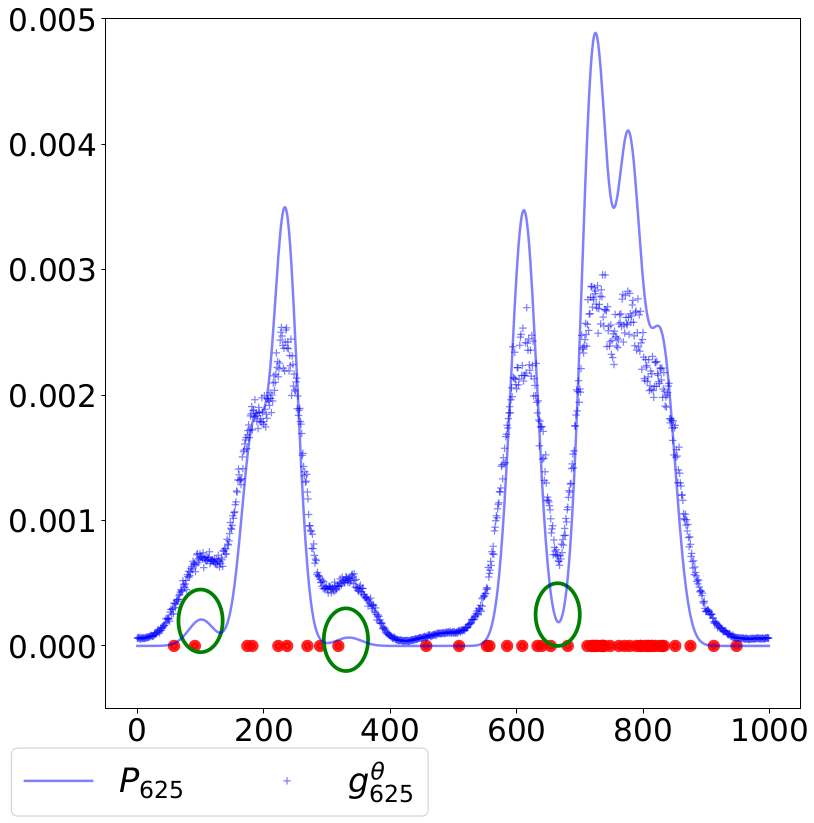}
    }\hfill
    \subcaptionbox{Negatives sampled with $B(E=G^\theta, D_i=1/P_i, T=0.75)$. With this temperature, we have an ideal density estimation. \label{ideal_temp}}
    {
        \includegraphics[width=0.31\linewidth]{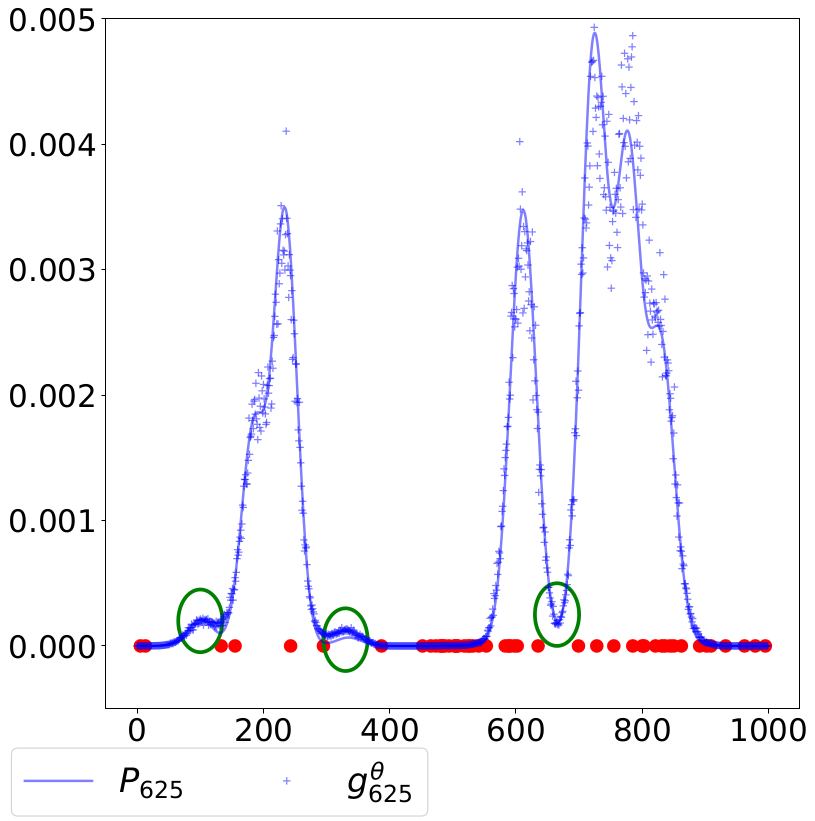}
    }\hfill 
    \subcaptionbox{Negatives sampled with $B(E=G^\theta, D_i=1/P_i, T=1.5)$. Very unlikely data is sampled; it leads to a poor estimation. \label{high_temp}}
    {
        \includegraphics[width=0.31\linewidth]{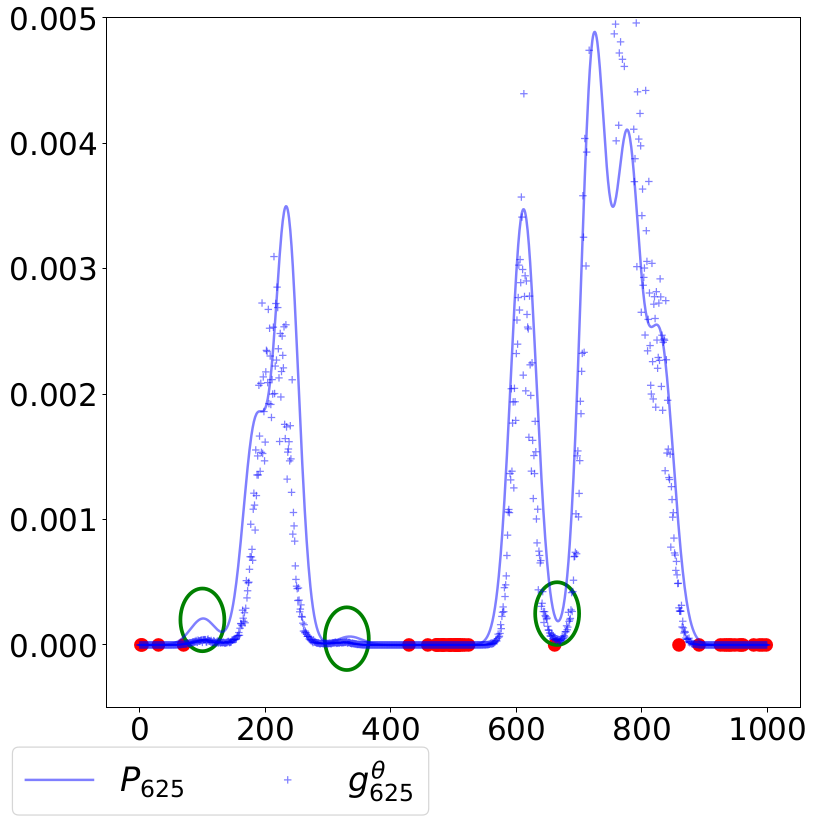}
    }
    \caption{Performance of the Boltzmann distribution negative sampling for different temperatures on a mixture of Gaussians. Difference in density estimation are highlighted in the green zones.}
    \label{fig:negative_samples_examples}
\end{figure*}

\begin{figure*}
    \centering
    \subcaptionbox{Benchmarking the temperature when the negative distribution has the following formulation $B(E=G^\theta, D_i=1/P_i, T^\star=0.75)$\label{benchmark_true_dist}}
    {   
        \includegraphics[width=0.31\linewidth]{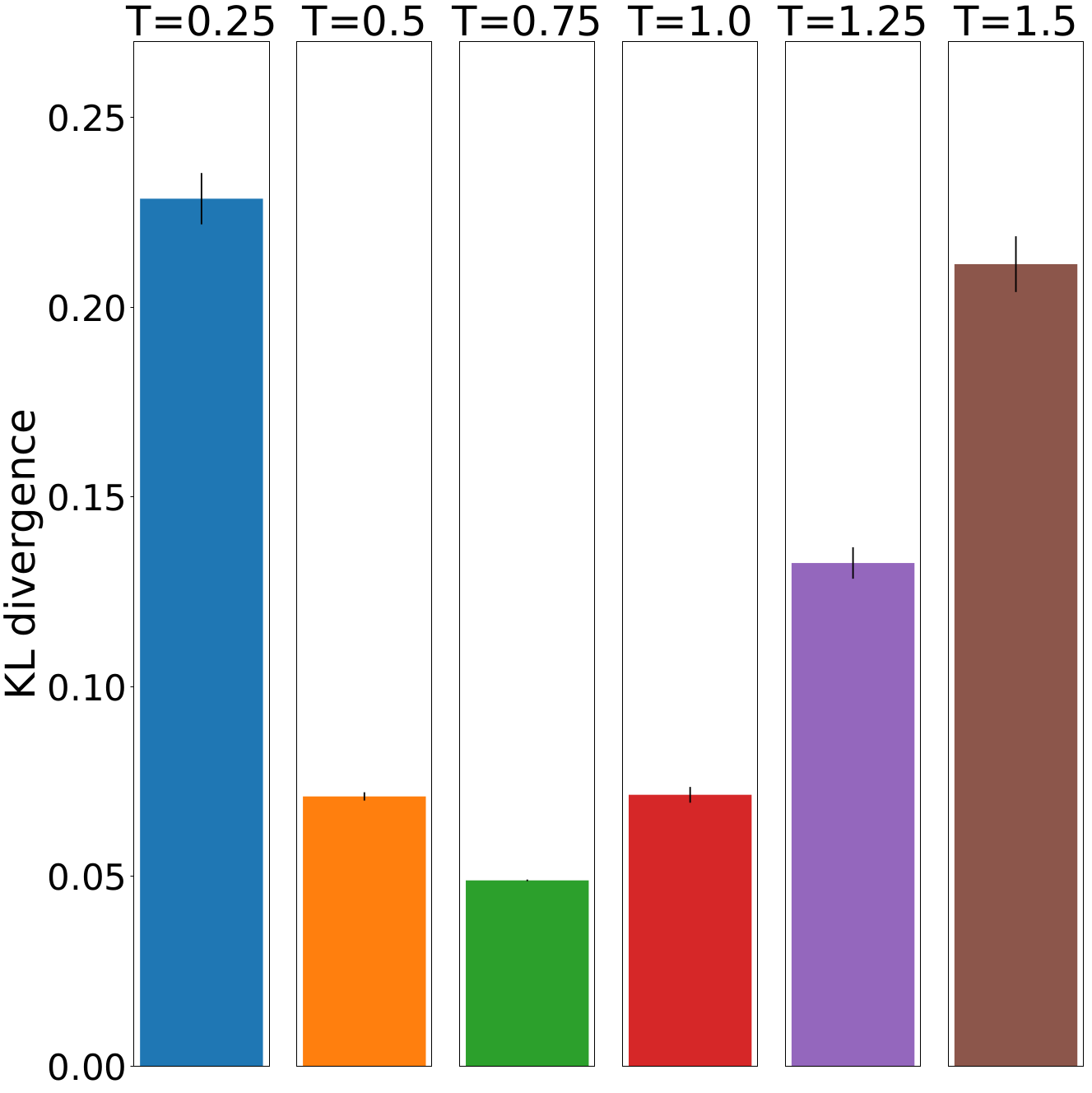}
    }\hfill
    \subcaptionbox{Benchmarking the temperature when the negative distribution has the following formulation $B(E=G^\theta, D_i=U(J), T^\star=5.0)$\label{benchmark_unif_dist}}
    {   
        \includegraphics[width=0.31\linewidth]{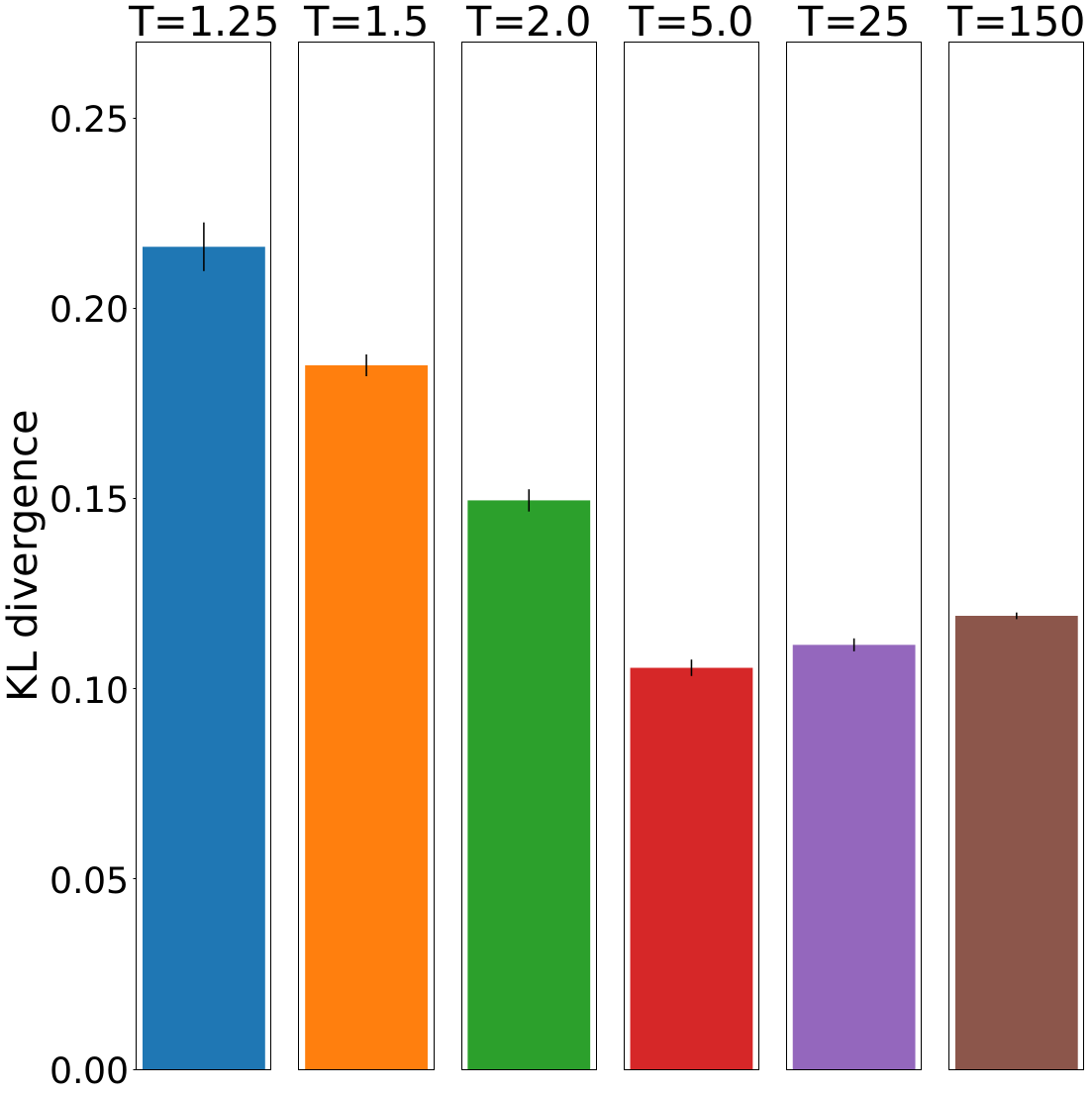}
    }\hfill
    \subcaptionbox{Benchmarking the temperature when the negative distribution has the following formulation $B(E=G^\theta, D=Pop(J), T^\star=7.5)$.\label{benchmark_pop_dist}}
    {   
        \includegraphics[width=0.31\linewidth]{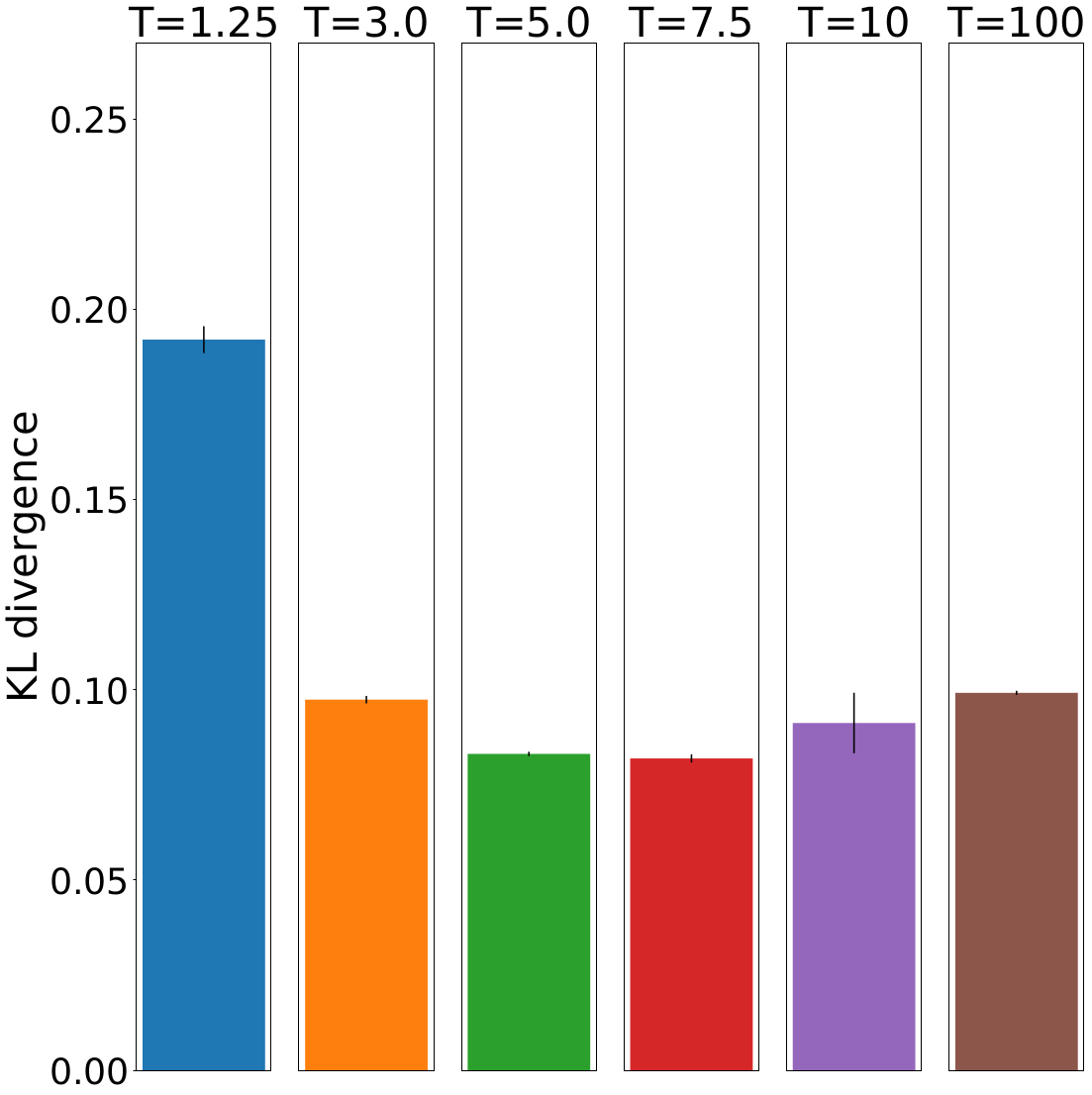}
    }
    \caption{Influence of the Temperature parameter $T$ on two degeneracy distributions on a mixture of Gaussians.}
    \label{fig:temperature_benchmarking}
\end{figure*}

Indeed, knowing the gradient when the number of samples goes to $\infty$ (Eq. \ref{eq:consistency_grad_NSP_loss}), we now define \textbf{a Boltzmann sampling distribution} for the \textit{Relaxed Softmax} loss: it is function of the energy state $E = G^\theta$, a given degeneracy distribution $D_i$ and a temperature parameter $T$, and has the following formulation:
\begin{equation}\label{eq:boltzmann_distribution}
    \forall j \in J, \ \ Q_i(j) = B_{(G^\theta, D_i, T)}(j) = \frac{D_i(j) \times \exp(G^\theta(i,j)/T)}{\sum D_i \times \exp(G^\theta/T)}
\end{equation}

Under this formulation of the \textit{Relaxed Softmax}, we notice that: 
\begin{itemize}
    \item This formulation helps us understand the model behavior: the RS loss (Eq. \ref{eq:negatively_sampled_softmax}) with the Boltzmann distribution $B_{(G^\theta, D_i, T)}$ (Eq. \ref{eq:boltzmann_distribution}) is simply the Monte-Carlo approximation of the gradient at Eq. \ref{eq:consistency_grad_NSP_loss} with $B = B_{(G^\theta, D_i, \frac{T+1}{T})}$.
    \item The RS loss generalizes Softmax: the softmax is a Boltzmann distribution with the specific parameters $D_i = U(J)$ ($U(J)$ is the uniform distribution) and $T = 1$. When using $D_i = U(J)$ and $T = \infty$, the RS loss is the Monte-Carlo approximation of the softmax cross-entropy gradient. One can cover many other distributions, playing with both $D_i$ and $T$.
\end{itemize}
 
Inspired from active learning and PU learning literature, we argue that the negative sampling scheme will hugely impact the performance of the model. Ideally, one would like to sample points that are misclassified by the model i.e. whose probability of being positive is overly estimated. Therefore, an optimal negative sampling distribution should take into account both how unlikely the data point is to be coming from the true generative process $P$ but also its estimated likelihood under our current model. This is exactly what is trying to approximate our proposed formulation: the energy state of each target item $G_i^\theta(j)$ is re-weighted by a distribution $D_i(j)$ that should emphasize the \textit{negativity} of each target. The temperature is designed to \textit{calibrate} the sampling distribution.

\subsubsection{Choosing the temperature parameter}

In the proposed sampling scheme, the temperature parameter is capital as it controls the trade-off between putting a stress on the energy state $E$ or the degeneracy distribution $D_i$. From an optimization standpoint, with low temperatures, the Boltzmann formulation will put more mass on samples $j \in J$ with high estimated scores $G_i^\theta(j)$ and explore less. On the other hand, with high temperatures, this Boltzmann formulation might focus too much on data points far way from the decision boundary that do not convey meaningful information:
\begin{itemize}
    \item When $T \to 0$, we have that $B_{(G^\theta, D_i, T)} \overset{\mathcal{L}}{\to} \delta(j^\star)$ where $j^\star = \underset{j \in J}{argmax} \ g_i^\theta(j)$. In this case, the negative chosen is the data point with highest estimated probability. We expect this model to behave poorly as it will lack exploration. 
    \item When $T \to \infty$, we have that $B_{(G^\theta, D_i, T)} \overset{\mathcal{L}}{\to} D_i$. In this case, we are guaranteed to sample very unlikely data that might not be insightful depending on the degeneracy distribution.
\end{itemize}
Unfortunately, we do not have any heuristics to find an optimal temperature parameter. It is clear that this ideal $T^\star$ will be a function of both the degeneracy function used but also the task tackled. In the next subsection, we analyze the temperature's impact on synthetic datasets on the density estimation task. Later on, in Section \ref{section:experiments}, we will see that for the same degeneracy distribution, optimizing for different tasks such as Mean Percentile Rank or test likelihood maximization leads to different ideal temperature parameters.

\subsubsection{Example on a toy dataset}

To better understand the behavior of our negative sampling distribution and the impact of the temperature parameter, we illustrate our approach on a synthetic dataset. We model a joint probability on the discrete set $I \times J$ with a mixture of Gaussians with 50 components. In this example, we have $card(I) = card(J) = 1000$. From this, we train our model $G^\theta$ to learn the conditional probability distributions. To do a comparison of the influence of the degeneracy distribution on the performance of the model, we bench three different degeneracy distributions: $D_i = U(J)$ the uniform distribution on the set of target items, $D_i = Pop(J)$ the popularity distribution on the set of target items, and finally, as we know $P$ in this synthetic dataset, we also bench $D_i(j) = 1/P_i(j)$. It is interesting to note that the last proposition is a context-based distribution that concentrates relatively more on data unlikely to be positive.

Figure \ref{fig:negative_samples_examples} draws attention to examples of negatives sampled conditionally to a specific context item $i \in I$ (in this example, $i=625$) for three different temperature parameters in the case where the degeneracy distribution is $D_i(j) = 1/P_i(j)$. Figure \ref{low_temp} stresses that with low temperatures, we are more likely to sample points with high estimated likelihoods. However, when the underlying generative process is easy to learn as it is the case in these synthetic datasets, the model ends up sampling positive data and misclassifies it as negative, deteriorating the performance. On the opposite, with high temperatures, one only samples very unlikely data points that do not bring meaningful information for the model as shown in Figure \ref{high_temp}: theses points are far from the decision boundary and the model has already classified them as negatives. We see from Figure \ref{ideal_temp} the existence of an optimal temperature parameter, in this case $T^\star = 0.75$, that achieves this interesting trade-off between \textbf{exploration} and sampling \textbf{close to the positives}. 

We now bench the temperature parameter with the three presented degeneracy distributions on the task of density estimation. Figure \ref{fig:temperature_benchmarking} reports the Kullback-Leibler divergence between the true joint distribution $P$ and an estimated one: $Pop(I) \times g_i^\theta$ where $Pop(I)$ refers to the popularity distribution on the context item. Figure \ref{fig:temperature_benchmarking} highlights the existence of \textbf{an optimal temperature parameter} for the three different degeneracy distributions. Interestingly, we can notice that for negative distributions that concentrate more on unlikely data, the optimal temperature parameter $T^\star$ is reached at lower values. Indeed, in these cases, one needs smaller values of $T$ to counter the effect of the degeneracy distribution. Finally, in the specific case when $D_i = 1/P_i(j)$, Figure \ref{benchmark_true_dist}, the model reaches a significantly better optimum. The newly proposed Boltzmann formulation takes therefore advantage of a negative sampling distribution that concentrates more on very unlikely data.

\section{Experiments}\label{section:experiments}

We now evaluate the proposed negative sampling methods on the tasks of density estimation, word similarities and analogies, and next-item prediction. All experiments below fall into the case of one-class problem where data points are sampled from a true generative process only known in the synthetic case. Our proposed method moves the problem to PU learning by artificially sampling unlabeled data points. To aid the reproducibility, we will release the source code shortly. In the following experiments, we used the following baselines: Maximum Likelihood Estimation training (MLE\footnote{\url{https://www.tensorflow.org/api_docs/python/tf/nn/softmax_cross_entropy_with_logits}}), Sampled Softmax with popularity sampling (SS\footnote{\url{https://www.tensorflow.org/api_docs/python/tf/nn/sampled_softmax_loss}}). Binary Cross-Entropy loss with popularity sampling (BCE\footnote{\url{https://www.tensorflow.org/api_docs/python/tf/nn/sigmoid_cross_entropy_with_logits}}) was not reported because, in our experiments, it lead to worse performances. We compared these baselines with the newly proposed \textit{Relaxed Softmax} (RS) loss used with a \textit{Boltzmann Sampling}. 

\subsection{The RS loss, a regularization scheme}

\begin{figure}
    \subcaptionbox{Averaged $KL(\hat{P_i}||g_i^\theta)$ among $i \in D^\star$. MLE has better training performances (Eq \ref{eq:softmax_gradient}). \label{batch_size_emp}}
    {   
        \includegraphics[width=0.47\linewidth]{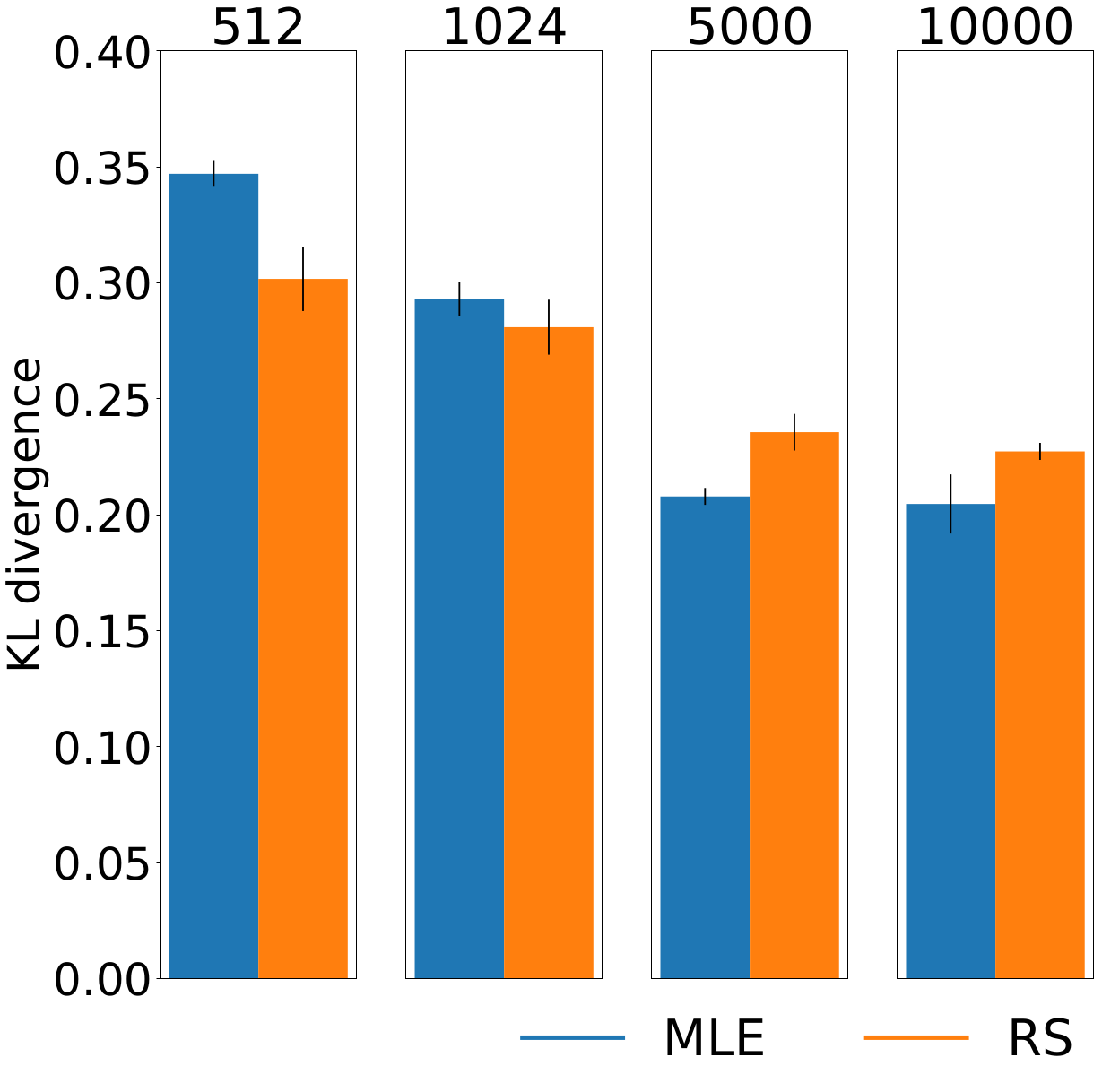}
    }\hfill
    \subcaptionbox{Averaged $KL(P_i||g_i^\theta)$ among $i \in D^\star$. The RS loss has better generalization properties. \label{batch_size_true}}
    {   
        \includegraphics[width=0.47\linewidth]{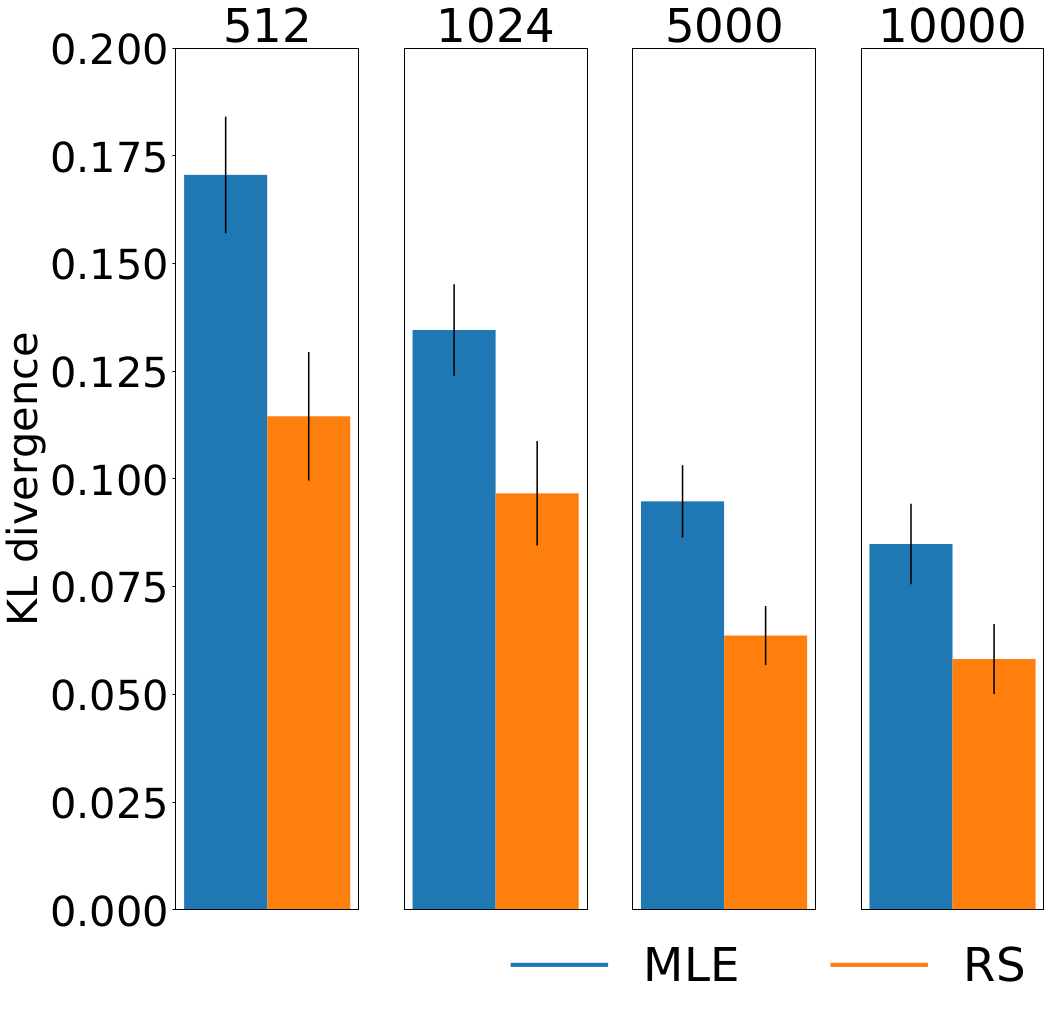}
    } 
    \caption{Influence of \textbf{the batch size} when tackling the task of density estimation. Training set size = 1,000,000.}
    \label{fig:mini_batch_properties}
\end{figure}

\begin{figure}
    \subcaptionbox{Averaged $KL(\hat{P_i}||g_i^\theta)$ among $i \in D^\star$. Comparable training performances between MLE and RS.}
    {   
        \includegraphics[width=0.47\linewidth]{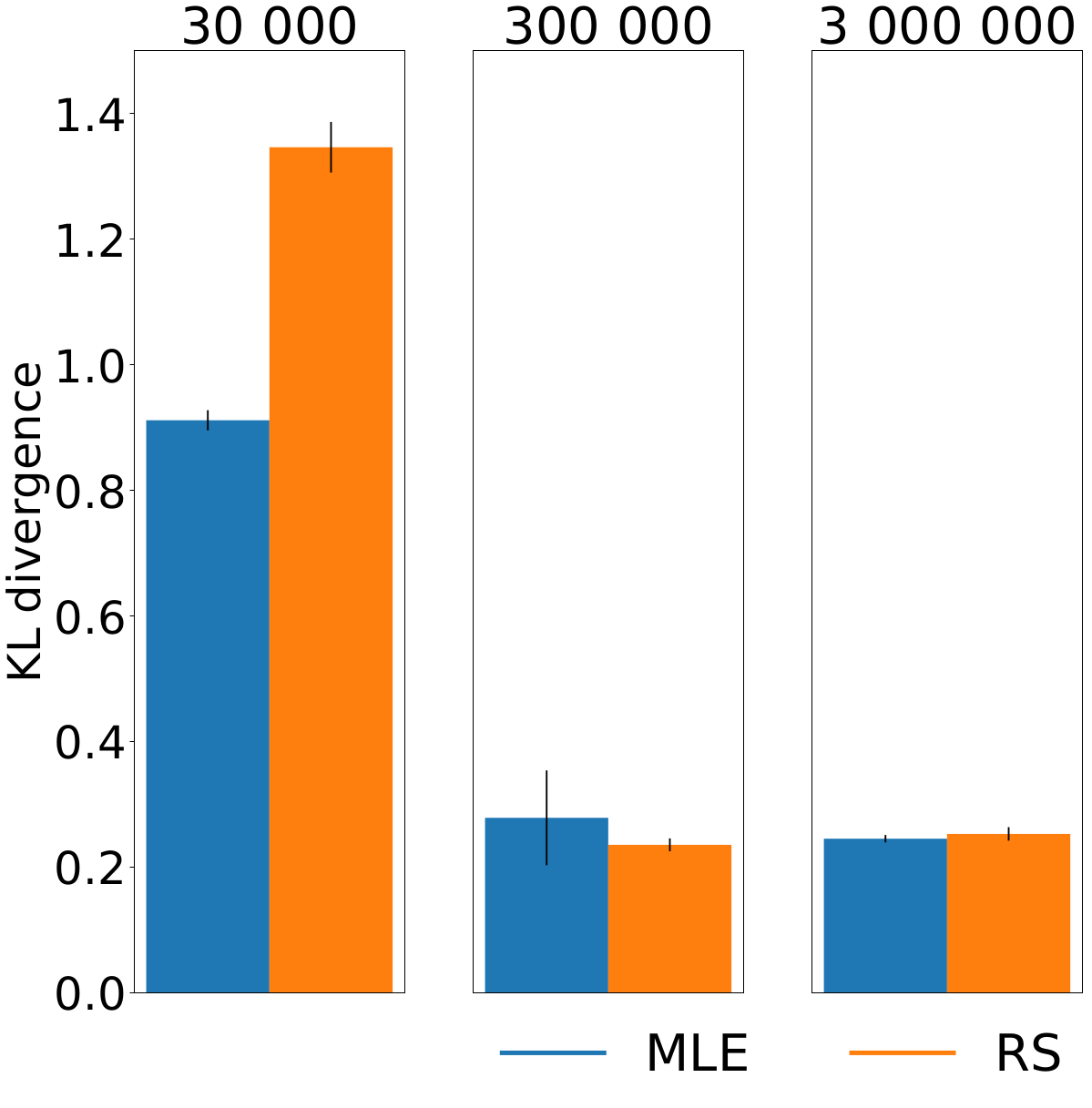}
    }\hfill
    \subcaptionbox{Averaged $KL(\hat{P_i}||g_i^\theta)$ among $i \in D^\star$. MLE generalization improves with the training size. \label{training_size_true}}
    {   
        \includegraphics[width=0.47\linewidth]{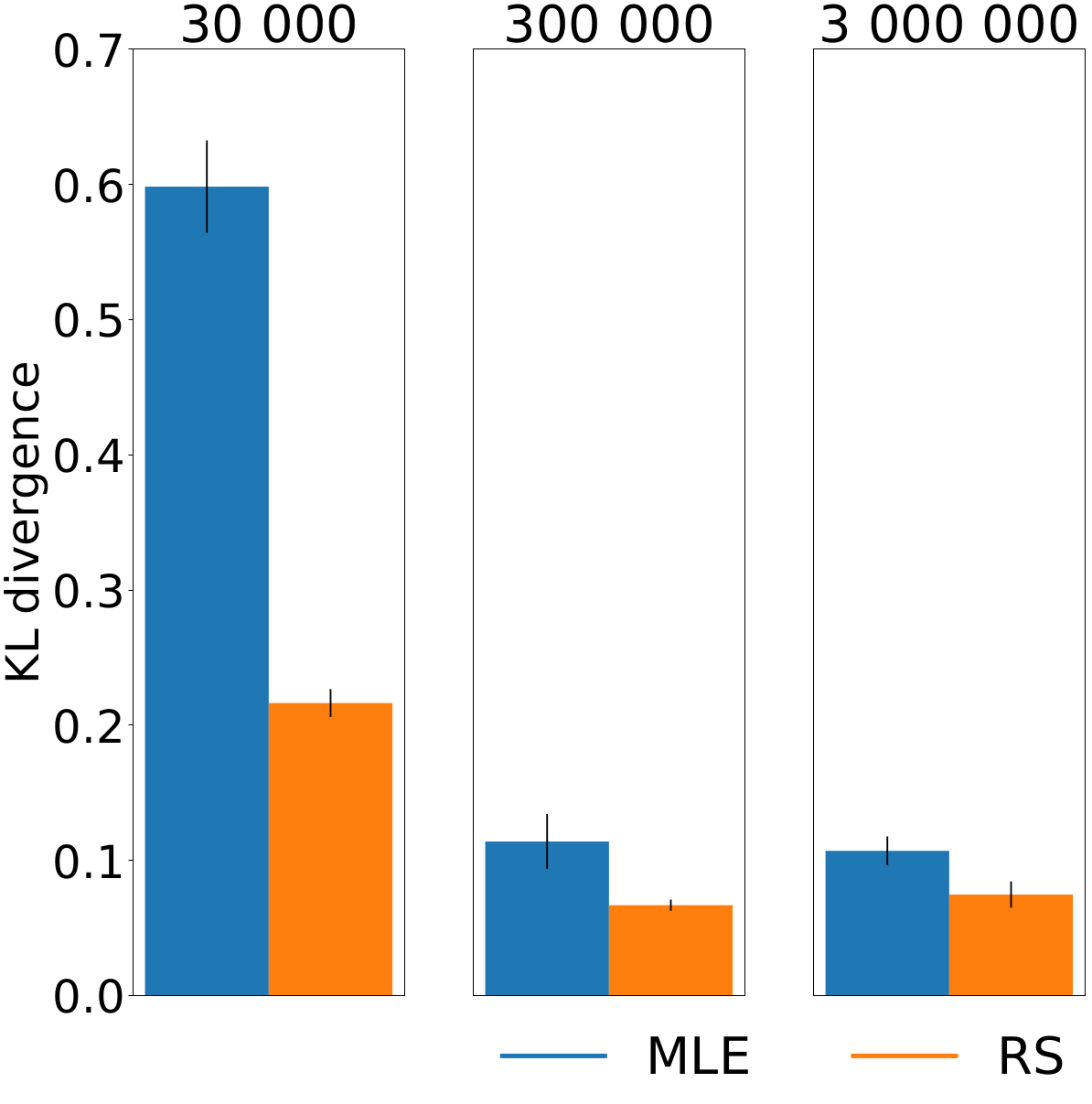}
    }
    \caption{Influence of \textbf{the training set size} when tackling the task of density estimation. Batch size = 2,000.}
    \label{fig:regularization_properties}
\end{figure}

To better understand the impact of the RS loss in the density estimation task, we first study the case of synthetic datasets. Similarly, to the toy dataset example in Section \ref{section:approach}, we model the joint distribution on $I \times J$ with a discretized mixture of Gaussians $P$. Doing so, we can have access to the true distribution and also control the complexity of the underlying data. In Figure \ref{fig:mini_batch_properties} and \ref{fig:regularization_properties}, we compare the performance of the standard MLE training with the newly defined RS loss when the negative distribution is $B(E=G^\theta, D_i=1/P_i, T=0.75)$ and the number of negatives sampled is 5. To stress the fact that the RS loss acts as regularizer, we want to show it improves generalization properties of the model and therefore report both training and test performances. We focus on learning conditional distributions of the context items only seen during training and report $KL(\hat{P_i}||g_i^\theta)$ and $KL(P_i||g_i^\theta)$ averaged over all $i \in D^\star$. Figure \ref{fig:mini_batch_properties} focuses on the influence of the batch size on the performances while Figure \ref{fig:regularization_properties} highlights the impact of the training set size.

Interestingly, we can see in Figure \ref{batch_size_emp} that MLE performs better on learning the empirical conditional distribution when dealing with large mini-batches. This makes sense as full-batch MLE is specifically minimizing the KL between the empirical distribution $\hat{P_i}$ and the model $g_i^\theta$. However, compared to the RS loss, we can see that the true conditional distribution $P_i$ is poorly fitted especially when the number of training samples is small, Figure \ref{training_size_true}. As the number of training points increases in the dataset, the performance of MLE training matches the performance of the RS loss in terms of generalization. We can therefore interpret the \textit{Relaxed Softmax} loss as a regularization scheme that can be efficient with the right prior.

\subsection{The next-item prediction task}

In practice, one does not have access to the true distribution. We want to see how our method performs on the task of next-item prediction without using any prior knowledge of the true distribution. Therefore, the degeneracy distributions used are the uniform and the popularity-based distribution. \textbf{For all of the following experiments, both on synthetic and real-world datasets, we compared six different training methods:}
\begin{itemize}
    \item MLE (Eq. \ref{eq:softmax_loss})
    \item Sampled Softmax (SS) with popularity sampling with 50 negative samples to approximate Eq. \ref{eq:softmax_gradient}.
    \item RS (Eq. \ref{eq:negatively_sampled_softmax}) with \textbf{Uniform Sampling (US)}.
    \item RS (Eq. \ref{eq:negatively_sampled_softmax}) with \textbf{Popularity-based Sampling (PS)}.
    \item RS (Eq. \ref{eq:negatively_sampled_softmax}) with \textbf{Uniform Botlzmann Sampling (UBS)}, $Q_i = B(E = G^\theta, D = U(J), T)$.
    \item RS (Eq. \ref{eq:negatively_sampled_softmax}) with \textbf{Popularity Boltzmann sampling (PBS)}, $Q_i = B(E = G^\theta, D = Pop(J), T)$.
\end{itemize}
When using the RS loss, across all experiments, we set the number of negative samples to 5 which was found to be optimal after grid search and cross validation. Besides, when using the Boltzmann sampling on real-world datasets, the temperature parameter was benched with the following grid: [0.5, 1, 3, 6, 12, 36].

\subsubsection{Benchmarking on synthetic datasets}

\begin{figure}
    \includegraphics[width=0.90\columnwidth]{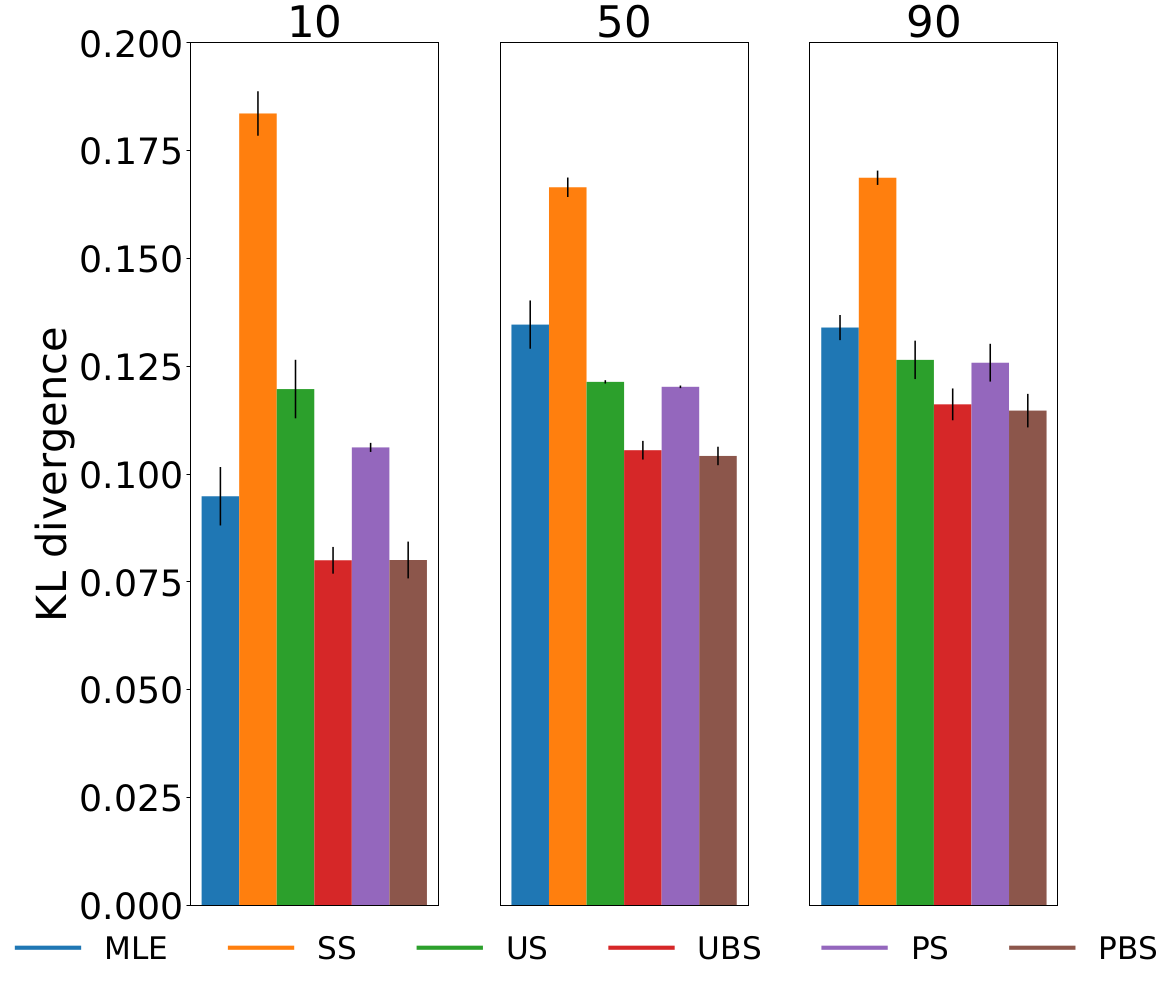}
    \caption{Comparing RS loss with different negative sampling schemes on three synthetic datasets made with mixtures of Gaussians (Number of Components=10, 50, 90). Our methods, Uniform Boltzmann Sampling (UBS) with $T^\star=5.0$ and Popularity-Based Sampling (PBS) with $T^\star=7.5$ outperforms the baselines on all three datasets. Black error-bars represent variance over 10 runs.}
    \label{fig:comparison_synthetic_datasets}
\end{figure}
Similarly to the previous subsection, we model $P$ with a 2-dimensional mixtures of Gaussians. Then, we generate three different datasets with respectively 10, 50 and 90 components to vary the complexity of the task. We bench the performance of the Boltzmann formulation on density estimation without any previous knowledge of the true distribution. On these synthetic datasets, the training set size was 300,000 data points and the batch size was 512 (as it was the optimal batch size for MLE for this training set size). We can see in Figure \ref{fig:comparison_synthetic_datasets}, that the Boltzmann formulation brings a significant uplift for both the uniform and popularity distributions on all three datasets. On the three different datasets, our RS loss with \textit{Boltzmann sampling} beats the baselines.

\subsubsection{Word Similarity and Analogy tasks}

\begin{table}
  \centering 
  \begin{tabular}{l|c|c|c|c|c|c|c}
  Method & MLE & SS & US & UBS & PS & PBS \\
  \hline\hline
  Similarities \\
  \hline
  WS353 & 0.46 & 0.44 & 0.38 & 0.50 & 0.52 & \textbf{0.55} \\
  SimLex999 & 0.13 & \textbf{0.26} & 0.25 & \textbf{0.25} & 0.21 & \textbf{0.25} \\
  Rare Word & 0.30 & 0.24 & 0.18 & 0.30 & \textbf{0.35} & \textbf{0.35} \\
  Bruni men & 0.42 & 0.38 & 0.33 & 0.49 & \textbf{0.52} & \textbf{0.52} \\
  Mturk & 0.47 & 0.40 & 0.34 & 0.61 & 0.53 & \textbf{0.64}\\
  
  \hline
  Semantic Analogies \\
  \hline
  Precision@1 & 15.4 & 11.7 & 13.5 & 13.0 & 13.3 & \textbf{15.9} \\
  Precision@5 & \textbf{29.5} & 24.8 & 21.9 & 26.6 & 25.0 & 24.5 \\
  Precision@15 & 39.7 & 33.0 & 31.3 & \textbf{40.0} & \textbf{39.5} & \textbf{40.5} \\
  \hline
  Syntactic Analogies \\
  \hline
  Precision@1 & 2.8 & 2.0 & 1.8 & \textbf{4.0} & 2.1 & 3.2 \\
  Precision@5 & 13.2 & 12.7 & 11.2 & \textbf{16.8} & 14.8 & 16.1 \\
  Precision@15 & 20.0 & 17.9 & 16.5 & 25.1 & 23.6 & \textbf{26.1} \\
  \end{tabular}
  \caption{Both of our solutions, UBS and PBS outperforms other methods on the Similarity and Analogy task. UBS (Uniform Boltzmann Sampling) is used with $T^\star=0.5$ and PBS (Popularity Boltzmann Sampling) with $T^\star=1.0$. Results with statistically significant improvements are highlighted.}
  \label{table:similarity_and_analogy_task_quantitatif}
\end{table}

Learning useful representations for words has been a largely studied field in Machine Learning and is of primal importance in both language modeling and recommendation where one is interested in computing similarities between products or movies. Authors in \cite{w2v} proposed to test their word embeddings on both similarity and analogy tasks to measure the expresiveness of the learned embedding space. Both tasks have now largely been accepted by the Natural Language Processing community. To learn our word embeddings, we trained a Word2vec model on the \textit{text8} dataset, a textual data excerpt from Wikipedia\footnote{\url{http://mattmahoney.net/dc/}}, and used the 15,000 most-common words for our experiments. On all experiments, we set the window size to 3. Then, we used different negative sampling distributions and benched them on both tasks of word similarities and analogies. All results are shown in Table~\ref{table:similarity_and_analogy_task_quantitatif}. For the similarity task, we use five public datasets that list pairs of words and their similarity scores such as : WordSimilarity353 \footnote{\url{http://www.cs.technion.ac.il/~gabr/resources/data/wordsim353/}}, SimLex-999 \footnote{\url{https://fh295.github.io/simlex.html}}, Rare Word \footnote{\url{https://nlp.stanford.edu/~lmthang/morphoNLM/}}, Men Test collection \footnote{\url{https://staff.fnwi.uva.nl/e.bruni/MEN}} and the MTurk dataset \footnote{\url{http://citeseerx.ist.psu.edu/viewdoc/download?doi=10.1.1.205.8607&rep=rep1&type=pdf}}. We compute the Pearson correlation between these human-annotated scores and the similarity scores output by the model. Analogy tasks are also used to compare embedding structures. With the pair-based analogy task used here, given word $i$ and its analogy $j$, one has to find, given a third word $i'$, its analogy $j'$. We consider two types of analogy tasks: \textit{semantic} and \textit{syntactic} analogies. We used the Google analogy test set \footnote{\url{http://download.tensorflow.org/data/questions-words.txt}} developed by~\cite{w2v} for our experiments. Semantic analogies test the robustness of word embeddings to transformations such as "man" to "woman", while syntactic analogies focus on transformations such as "run" to "runs". We show quantitative results for the prec@1, prec@5 and prec@15 metrics in the Table \ref{table:similarity_and_analogy_task_quantitatif}.

The Boltzmann formulation brings an uplift for both similarity and analogy task. Optimal temperatures are lower compared with those of the task of density estimation on synthetic datasets: in this case, $T^\star = 0.5$ when $D_i = U(J)$ and $T^\star = 1$ when $D_i = Pop(J)$. In ranking tasks, one focuses on predicting correctly the top of the distribution, thus the need for \textit{harder negatives}.

\subsubsection{Next-Item prediction}

\begin{table}
  \centering
  \begin{tabular}{l|c|c|c|c|c|c}
  Method & MLE & SS & US & UBS - $T^\star$ & PS & PBS - $T^\star$ \\
  \hline\hline
  Likelihood \\
  \hline
  Mvls (e-4) & 5.1 & 4.5 & 6.0 & 6.0 - $\infty$ & \textbf{6.5} & \textbf{6.5 - $\infty$} \\
  Netflix (e-4) & \textbf{6.0} & 5.2 & 3.2 & 3.2 -$\infty$ & 3.0 & 3.0 - $\infty$ \\
  Text8 (e-2) & 1.65 & 1.7 & \textbf{1.9} & 1.9 - $\infty$ & 1.8 & 1.8 - $\infty$ \\
  \hline
  MPR \\
  \hline
  Mvls & 91.7 & 91.5 & 89.0 & \textbf{92.5 - 6} & 92.2 & 92.1 - 6 \\
  Netflix & 87.3 & 86.7 & 87.8 & \textbf{88.7 - 12} & 87.6 & \textbf{89.0 - 12} \\
  Text8 & 87.2 & 88.2 & 89.7 & \textbf{90.7 - 12} & 89.4 & \textbf{90.8 - 12} \\
  \hline
  Prec50 \\
  \hline
  Mvls & 4.9 & 5.5 & 6.1 & \textbf{7.0 - 6} & 6.2 & \textbf{7.3 - 6} \\
  Netflix & 3.7 & 3.9 & 4.2 & \textbf{5.1 - 6} & 4.1 & \textbf{5.5 - 6} \\
  Text8 & 31.9 & 31.2 & 31.5 & \textbf{34.2 - 6} & 31.5 & \textbf{34.5 - 6} \\
  \hline
  Prec15 \\
  \hline
  Mvls & 1.3 & 1.7 & 2.3 & 2.6 - 3 & 2.2 & \textbf{3.0 - 3} \\
  Netflix & 1.2 & 1.4 & 1.2 & 1.8 - 3 & 1.3 & 1.5 - 3 \\
  Text8 & 13.1 & 12.0 & 12.5 & \textbf{14.2 - 1} & 11.6 & \textbf{14.5 - 1} \\
  \hline
  Prec5 \\
  \hline
  Mvls & 0.4 & 0.5 & 0.7 & 0.7 - 3 & 0.6 & \textbf{1.0 - 3} \\
  Netflix & 0.4 & 0.4 & 0.5 & 0.7 - 3 & 0.5 & 0.5 - 3\\
  Text8 & 4.5 & 4.2 & 4.4 & \textbf{5.1 - 1} & 4.1 & 4.6 - 1 \\
  \end{tabular}
  \caption{Likelihood, MPR (in \%) and Precision (in \%) on the next-item prediction task. Confidence intervals are computed over 10 runs and results statistically better have been highlighted. On all ranking tasks (MPR and Prec@k), our Botlzmann formulation outperform the different baselines.
  }
  \label{table:results_real_data}
\end{table}

Finally, we want to compare the different methods on the task of the next-item prediction. We therefore performed experiments on three different public datasets: two movie recommendation datasets listing movies seen by users (the \textit{Movielens} and the \textit{Netflix} datasets), and the same text dataset excerpt from Wikipedia. 
For the movie datasets, we kept only movies ranked over 4 stars for the Movielens dataset and over 5 stars for the Netflix dataset, resulting in 17,128 and 17,770 movies for each dataset, respectively. From the user-to-item co-occurrence matrix we create an item-to-item co-occurrence matrix, where a pair is considered positive if both items are seen together in a user timeline. For the text dataset, we kept only the 15k most common words and considered a pair to be positive if both words co-occurred within a window of size 3. In order to learn relevant conditional probabilities, the sequential aspect of the data was kept and the co-occurence matrix was not made symmetrical. We use all of these datasets to perform an implicit matrix factorization, with access to only positive data.

We split each dataset so that 70\% is used for training, 10\% for validation, and 20\% for testing. In Table \ref{table:results_real_data}, we report the following metrics on the next-item prediction task:
\begin{itemize}
    \item Likelihood of pairs in the test set. Given the true generative process $P$ and the estimated output probability $g_i^\theta$, likelihood is calculated as follows:
     \begin{equation}
         \mathbb{E}_{(i,j) \sim P} \ \ g_i^\theta(j)
     \end{equation}
    \item Approximated Mean Percentile Rank: frequency with which positive target items $j \sim P_i$ are ranked over uniformly sampled negative target items $j' \sim U(J)$.
    \begin{equation}
        \mathbb{E}_{(i,j) \sim P} \ \ \mathbb{E}_{j' \sim U(J)} \ \ \mathbf{1}_{g_i^\theta(j) > g_i^\theta(j')}
    \end{equation}
    \item Precision@k : Frequency with which positive target items $j \sim P_i$ are ranked in the top k among all potential targets $j' \in J$. We report Prec@5, Prec@15 and Prec@50.
\end{itemize}
The likelihood metric is a density estimation metric whereas Approximated MPR and Precision@k are ranking metrics. Interestingly, Precision@k only focuses on the top of the distribution disregarding how well can we rank the tail. 

As we can see in Table \ref{table:results_real_data}, our proposed solution brings significant uplift from conventional methods when dealing with ranking metrics such as MPR and Prec@k. We can notice that when ranking metrics are focusing on the top of the distribution, as with Prec@5 and Prec@15 for example, the optimal temperatures $T^\star$ get smaller. Indeed, optimality for these metrics is reached when negatives are sampled really close to the positive data. In these specific cases, it does not matter to sample data likely to be positive as the task really focuses on the top items. Contrary to previous generic sampling schemes, the advantage of the Botlzmann formulation is that we can now fit the negative sampling distribution to the given task. 

Interestingly, on the task of density estimation on these real-world dataset, we can see that the Botlzmann formulation does not impact much the result. In this task, the optimal temperature parameter is $T^\star = \infty$ for both degeneracy distributions. Table \ref{table:results_real_data} stresses the importance of having a negative sampling distribution fit to the task. The Boltzmann formulation can prove itself efficient as it enables us to bench a variety of distributions. 

Finally, we can notice from the results that the uplifts in performance are larger when it comes to textual data. We argue that compare to language modeling, sequences of movies convey less information. Additional works will analyze the impact of our proposed solution on the user-item prediction task.

\section{Conclusions}

In this paper, we proposed a \textit{Relaxed Softmax} loss, a sampling-based loss that is more suited to the case of PU learning. With a newly defined \textit{Boltzmann Sampling}, where one can fit the negative sampling distribution to a specific task, we have shown significant improvements over generic sampling methods on tasks such as density estimation, language modeling and next-item prediction.

In future work, we will investigate the effectiveness of this training procedure on more complex models that can also be trained with negative sampling, such as CNN, RNN or \textit{Transformer} like models \cite{vaswani2017attention}. Models with more capacity than \textit{Word2Vec} could benefit even more from using our dynamic context-based sampling. 

Building upon the work in \cite{blei2003latent, yang2018breaking} and assuming that language is multi-topic, we would like to develop a new sampling scheme that also takes into account the topic of the target item. 

Finally, we have shown the existence of an optimal temperature parameter for different degeneracy distributions and tasks. However, in all our experiments, this temperature was kept fixed for each run. Later research work should try and look for the advantages of an adaptive temperature during the training.

\newpage
\bibliography{adversarial_training}
\bibliographystyle{ACM-Reference-Format}

\end{document}